\documentclass[sigconf]{acmart}
\renewcommand\footnotetextcopyrightpermission[1]{} 
\usepackage{graphicx}
\usepackage{subfig}
\usepackage{amsmath}
\usepackage{booktabs}
\usepackage{amsthm}
\usepackage{amssymb}
\usepackage[skip=2pt]{caption}
\AtBeginDocument{%
  \providecommand\BibTeX{{%
    \normalfont B\kern-0.5em{\scshape i\kern-0.25em b}\kern-0.8em\TeX}}}
\newcommand\blfootnote[1]{%
	\begingroup
	\renewcommand\thefootnote{}\footnote{#1}%
	\addtocounter{footnote}{-1}%
	\endgroup
}

\begin{document}

\title{Deep Stock Predictions}

\author{Akash Doshi}
\affiliation{%
}
\email{akashsdoshi@utexas.edu}

\author{Alexander Issa}
\affiliation{%
}
\email{alex.issa32@utexas.edu}

\author{Puneet Sachdeva}
\affiliation{%
}
\email{puneet\_sachdeva@utexas.edu}

\author{Sina Rafati}
\affiliation{%
}
\email{rafati@utexas.edu}

\author{Somnath Rakshit}
\affiliation{%
}
\email{somnath@utexas.edu}

\begin{abstract}
  Forecasting stock prices can be interpreted as a time series prediction problem, for which Long Short Term Memory (LSTM) neural networks are often used due to their architecture specifically built to solve such problems. In this paper, we consider the design of a trading strategy that performs portfolio optimization using the LSTM stock price prediction for four different companies. We then customize the loss function used to train the LSTM to increase the profit earned. Moreover, we propose a data driven approach for optimal selection of window length and multi-step prediction length, and consider the addition of analyst calls as technical indicators to a multi-stack Bidirectional LSTM strengthened by the addition of Attention units. We find the LSTM model with the customized loss function to have an improved performance in the training bot over a regressive baseline such as ARIMA, while the addition of analyst call does improve the performance for certain datasets.
\end{abstract}

\maketitle
\blfootnote{Author names listed in alphabetical order. All authors are from The University of Texas at Austin, Austin, TX, USA.}

\section{Introduction {\&} Related Work} \label{sec:intro}
Efficient functioning of stock markets requires market intermediaries who trade stocks for a short duration and keep the market liquid. Machine learning algorithms have been proposed to help such market intermediaries make better predictions for the short-term price movements \cite{li2018stock, gaurav2019towards, liu2019stock}. Both \cite{li2018stock} and \cite{liu2019stock} use LSTMs to predict the stock price. They perform various architectural modifications to improve their respective metrics. In particular, \cite{liu2019stock} used sparse auto-encoders with 1-D residual convolutional networks to denoise the data and improve the mean absolute percentage error (MAPE), while \cite{li2018stock} uses an Attention \cite{bahdanau2014neural} mechanism to improve the mean squared error (MSE) in stock price prediction. In \cite{gaurav2019towards}, they use a deep FLANN (functional link artificial neural network) architecture, which is similar to a feed-forward Neural Network (NN) with time-varying weights to predict the stock prices. \\
\newline
All the prior work has thus been focused on minimizing some metric that drives the predictions close to the real stock price. However, this does not imply that these predictions will yield the maximum profit. For instance, if the real stock price decreases, but the LSTM predicts a slight increase, it would be more detrimental than an LSTM prediction that had a higher MSE but forecasted a decrease. To more fully exploit this observation, we first train a standard Multi-Stack LSTM and feed its predictions to a trading bot designed as a linear optimization program. We then modify our loss function to optimize forecasting the correct trend and see its impact on the trading bot. In addition, we explore the effect of adding correlated time-series indicators to our data, and perform a data-driven optimization of the LSTM hyperparameters to point us towards the optimal trading strategy. 

\section{Dataset {\&} Data Pre-processing}
In this work, we predict future stock prices for four companies in the automobile industry with the dates for the stock price data given alongside - Ford (1983 to 2020), GM (1985 to 2020), Toyota (1980 to 2020) and Tesla (2010 to 2020). Daily stock prices, specifically the daily open, close, low, and high stock prices were taken from the Capital IQ database \cite{phillips2012s} by Compustat through a Wharton WRDS subscription from the University of Texas at Austin. We utilized the Mid price, which is computed as an average of the High and Low price. To enrich the feature space of our dataset, some basic features like n-day moving averages values were derived. All data was normalized using Standard Scalar\footnote{http://scikit-learn.org/stable/modules/generated/sklearn.preprocessing.StandardScaler.html} fit to the training set. The size of the training and testing set are shown in Table \ref{tab:dataset_size}.
\begin{table}[h]
	\caption{Training and Testing Set Sizes for Four Companies}
	\label{tab:dataset_size}
	\begin{tabular}{@{}lcc@{}}
		\toprule
		Company & Training Set Size & Testing Set Size \\ \midrule
		Ford    &     8072              &       1074           \\
		Tesla   &       1388       &         1078        \\
		Toyota  &        8071           &      1082            \\
		GM      &           7815        &             1077      \\ \bottomrule
	\end{tabular}
\end{table}

\section{Proposed Models}

\subsection{ARIMA}
In time series analysis, the auto-regressive integrated moving average (ARIMA) model\footnote{http://alkaline-ml.com/pmdarima/0.9.0/modules/generated/pyramid.arima.auto\_arima.html} is a generalization of an auto-regressive moving average (ARMA) model\footnote{https://en.wikipedia.org/wiki/Autoregressive\_integrated\_moving\_average}. ARIMA is a simple stochastic time series model that can capture complex relationships since it takes error terms and observation of lagged terms. The advantage of ARIMA over exponential moving averages is the fact that the ARIMA model aims to describe the auto-correlation in the data while the exponential moving average describes only the overall trend \cite{hyndman2018forecasting}. Given the goal is to estimate the price of the stock for a future day, it is not possible to use the feature values of the same day since they are not going to be available at actual interface time. The remedy for that is a derivation of statistics like mean and standard deviation of their lagged values. To that end, we have used three sets of lagged values by looking at 3 days, 7 days, and 30 days back. Considering the auto-regression (AR), integrated (I), moving average specifications of ARIMA, there are a set of parameters that needs to be set for the model. The parameters are P, the number of lag observations included in the model, d, the number of times that the raw observations are differenced, and q, the size of the moving average window \cite{magiya_2019}. In this study, we have used the Auto ARIMA, which is an automatic process by which these parameters can be chosen. The optimum ARIMA was estimated with 50 iterations subjective to lowest MSE by minimizing Akaike Information Critera (AIC) and Bayesian Information Criterion (BIC) . The optimum lag was found to be 5 with zero order of differencing which is obtained with a window length varying from 0 to 7.   
\subsection{Regular Multi-Stack LSTM}
\begin{figure}[b]
	\centering
	\includegraphics[height=5cm,width=7cm]{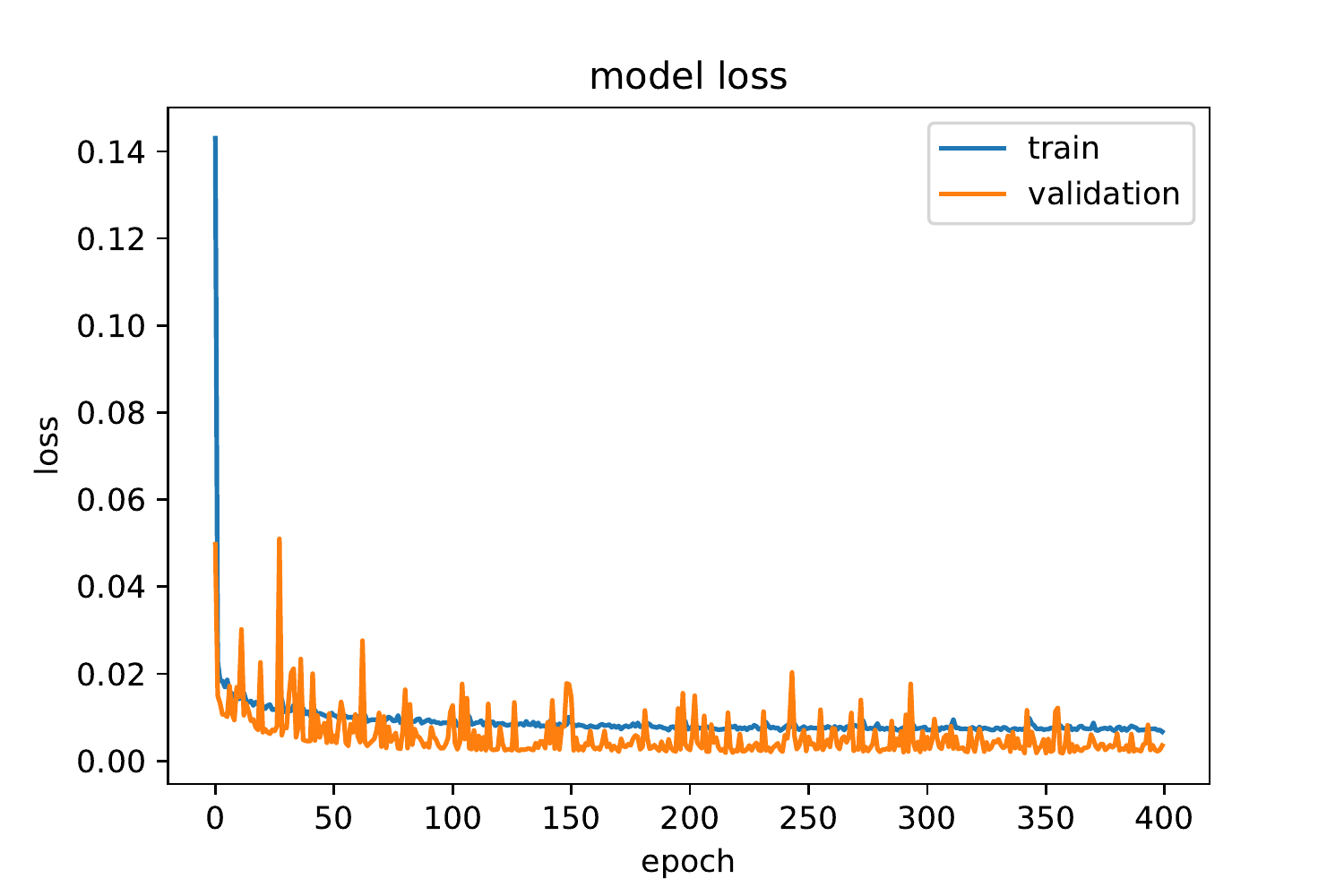}
	\caption{Loss Curve}
	\label{fig:loss_curve}
\end{figure}

A multi-stack LSTM neural network was used to predict the future stock price for the four companies of interest. Our LSTM architecture contains 4 LSTM layers alternated with 4 layer of 30\% Dropouts \cite{srivastava2014dropout}, added to prevent over-fitting. The output is then fed into a dense layer which gives the estimated stock price. The model uses the last 50 days of stock price as a window of time that is input into the model. The various hyperparameters used during training are summarized in Table \ref{tab:hyperparams}. 
\begin{table}[b]
	\caption{Various Hyperparameters used in this Work along with their Values}
	\label{tab:hyperparams}
	\begin{tabular}{@{}ll@{}}
		\toprule
		Hyperparameter & Value                                             \\ \midrule
		Learning Rate  & $5\times 10^{-3}$ \\
		Optimizer          & Adam \cite{kingma2014adam}       \\
		Batch size     & 256                                               \\
		Epochs         & 400                                               \\ 
		Loss Function & MSE \\\bottomrule
	\end{tabular}
\end{table}
The loss curve for Toyota is plotted in Figure \ref{fig:loss_curve}, using a 4:1 split between training and validation. As is evident, the training error converges at the end of the training period, as does the validation error, indicating the models can be used for prediction.

\subsection{Custom Loss for LSTM}
As discussed in Section \ref{sec:intro}, predicting the correct direction of movement of the stock price is more crucial to maximizing net worth. In other words, if $\mathbf{x}_t$ denotes the stock price, and $\mathbf{\hat{x}}_t$ denotes the predicted stock price at time $t$, then the actual change in price is $\mathbf{x}_{t+1} - \mathbf{x}_{t}$ while the predicted change in stock price is $\mathbf{\hat{x}}_{t+1} - \mathbf{x}_{t}$. If these two quantities are of opposite sign, the training procedure must penalize the prediction. Hence we modified the loss function as being non-zero only when the opposite signs condition is met:
\begin{equation}
    \mathcal{L}(\mathbf{\hat{x}}_{t+1}, \mathbf{x}_{t+1}) = ||\mathbf{\hat{x}}_{t+1} - \mathbf{x}_{t+1}||^2   \;\;\; (\mathbf{x}_{t+1} - \mathbf{x}_{t})(\mathbf{\hat{x}}_{t+1} - \mathbf{x}_{t}) < 0
\end{equation}

\section{Portfolio Optimization Bot} \label{sec:pob}
Consider a set of $N$ companies, whose associated stock prices at time $t$ are denoted by $\mathbf{x}_t \in \mathbb{R}^N$. At time $t$, the number of shares of an investor are denoted by $\mathbf{s}_t \in \mathbb{R}^N$. We assume that the investor is allowed to re-balance his portfolio on a daily basis, and uses the stock price prediction $\mathbf{\hat{x}}_{t+1}$ for time $t+1$ to design a new portfolio $\mathbf{s}_{t+1}$ that maximizes their expected percentage profit $\mathbf{s}_{t+1}^T (\mathbf{\hat{x}}_{t+1} - \mathbf{x}_t)/\mathbf{x}_t$. We also denote their wealth not invested at time t as $w_t$. Then we can formulate the daily portfolio optimization as a linear program: \begin{equation}
    \mathbf{s}^*_{t+1} = \underset{\mathbf{s}_{t+1} \in \mathbb{R}^N} {\mathrm{arg\ max\ }} \hspace{0.05 in} \mathbf{s}_{t+1}^T (\mathbf{\hat{x}}_{t+1} - \mathbf{x}_t)/\mathbf{x}_t,
\end{equation}
subject to a wealth re-balancing and positivity constraint:
\begin{equation}
    (\mathbf{s}_{t+1} - \mathbf{s}_t)^T \mathbf{x}_t + w_{t+1} - w_{t} = 0 \ \ \  \mathbf{s}_{t+1} \geq \mathbf{0} \ \  w_{t+1} \geq 0.
\end{equation}
The profit actually earned on day $t+1$ will be $\mathbf{s}^*_{t+1}(\mathbf{\hat{x}}_{t+1} - \mathbf{x}_t)$. We attempt to invest in the company with the maximum relative change in stock price, hence explaining the division by $\mathbf{x}_t$. As a baseline, we consider a trading strategy (HOLD) that invests equally in all companies and does not re-balance their portfolio.

\section{Results}
\subsection{LSTM vs. ARIMA }
\begin{figure*}[!t]
	\centering
	\subfloat[]
	{\includegraphics[width=4cm]{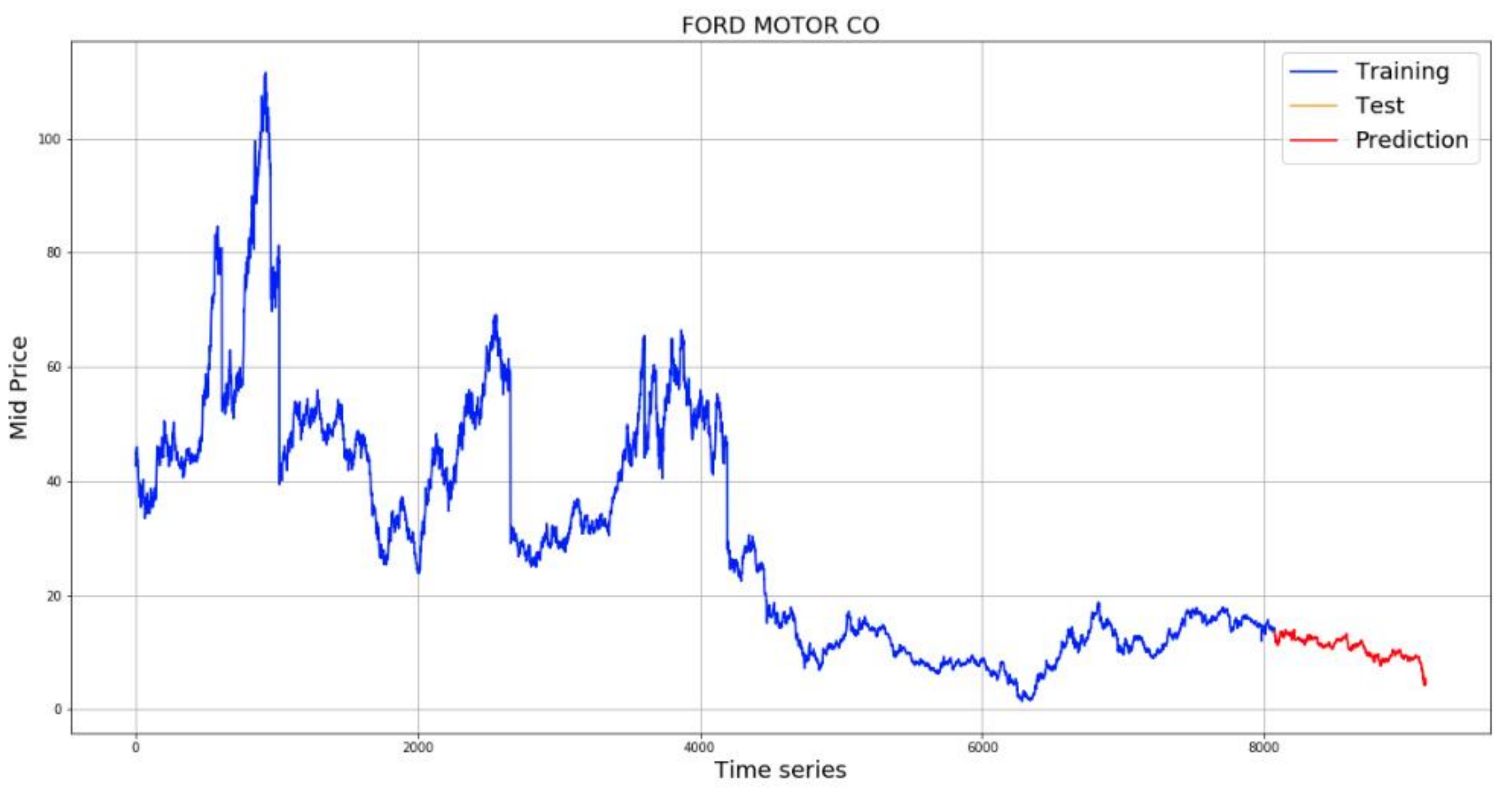}}
	\subfloat[]
	{\includegraphics[width=4cm]{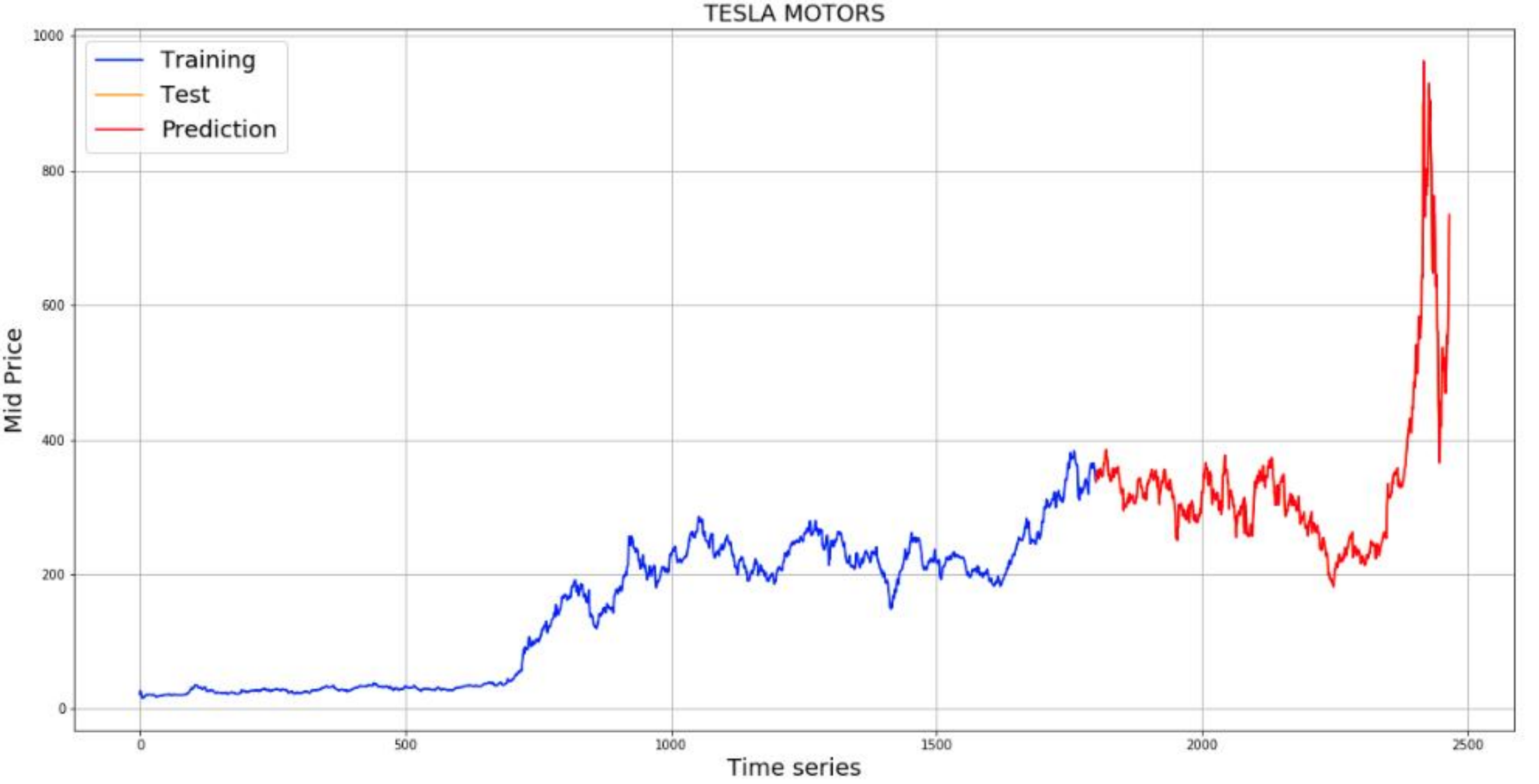}}
	\subfloat[]
	{\includegraphics[width=4cm]{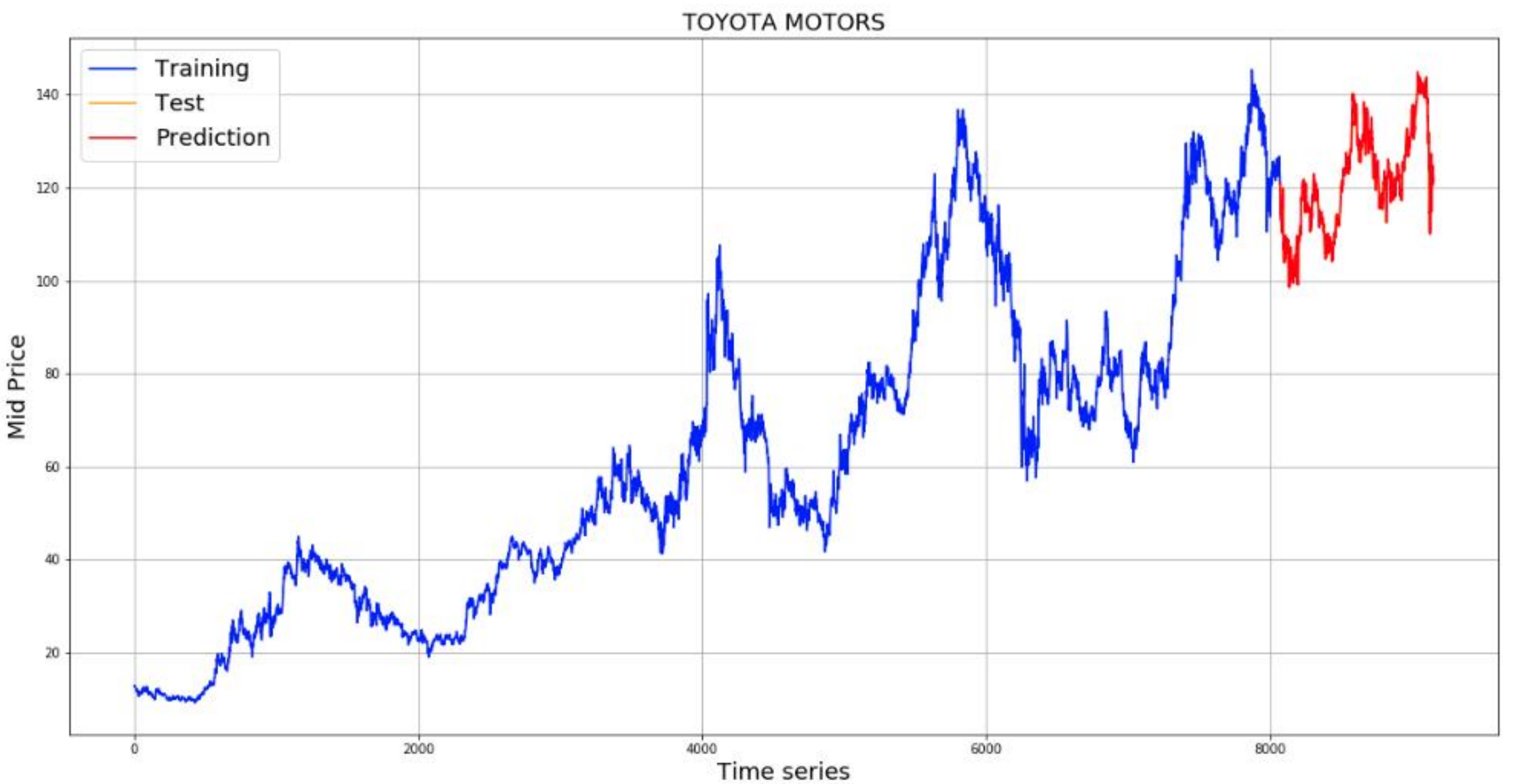}} 
	\subfloat[]
	{\includegraphics[width=4cm]{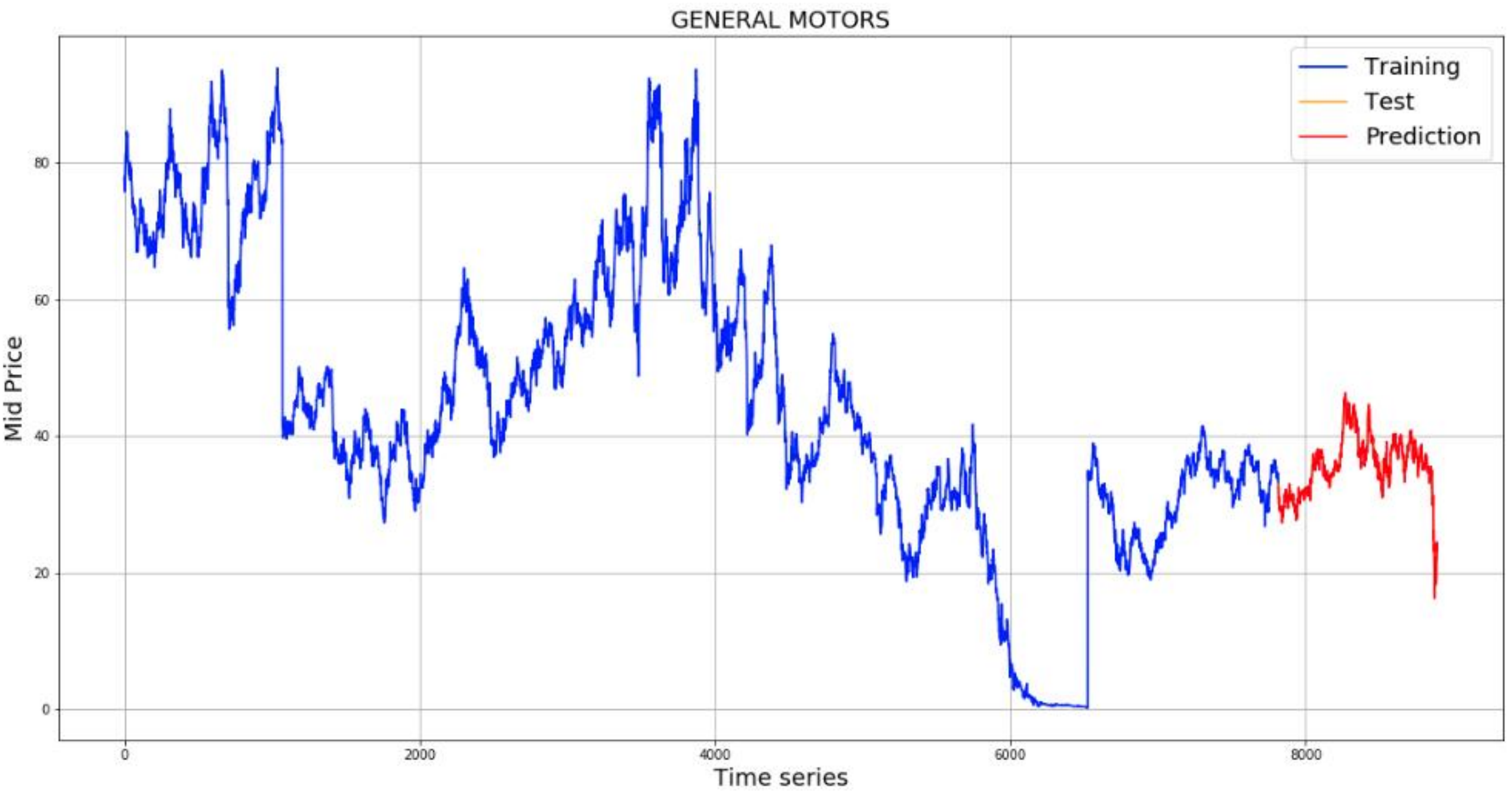}}
	
	\caption{Stock Price Predictions using ARIMA in case of Ford (a), Tesla (b), Toyota (c) and GM (d)}
	\label{img_grid_ARIMA}
\end{figure*}

\begin{figure*}
	\centering
	\subfloat[]
	{\includegraphics[width=4cm]{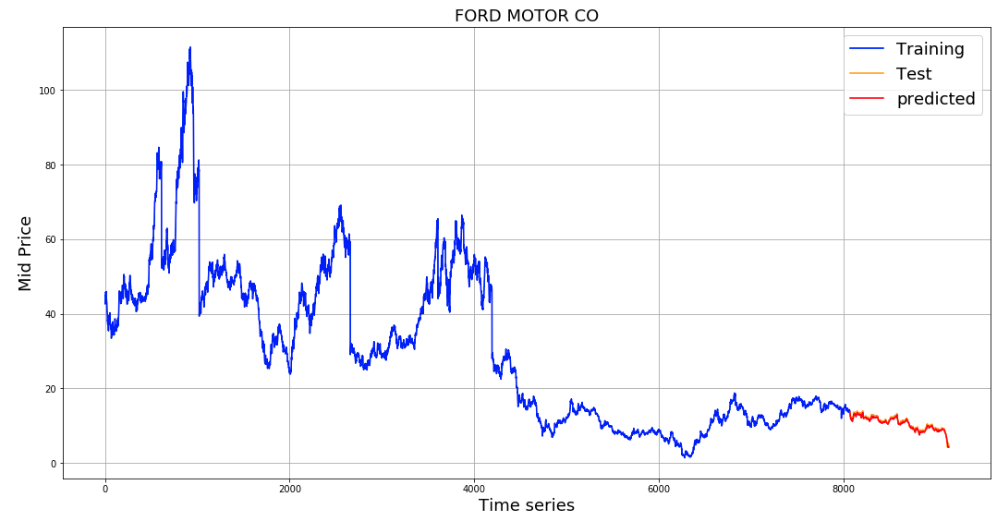}}
	\subfloat[]
	{\includegraphics[width=4cm]{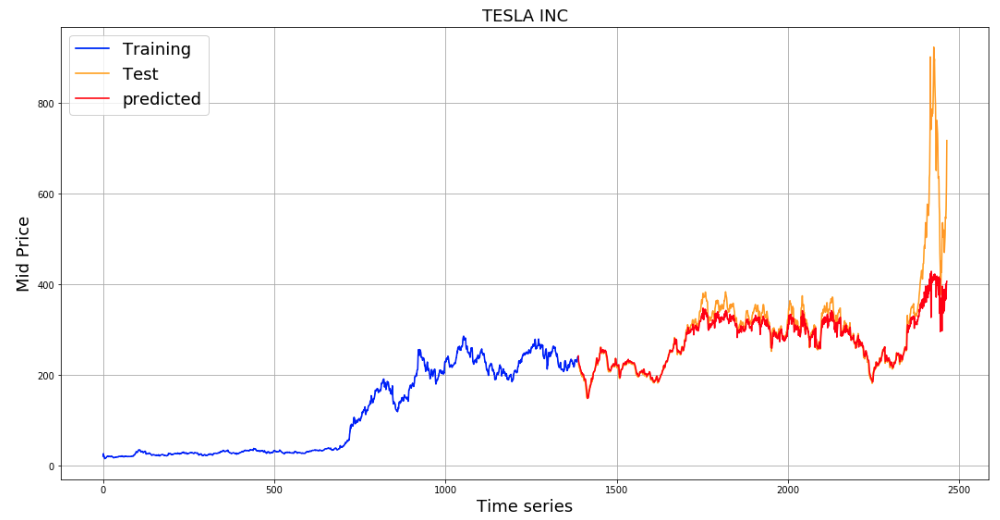}}
	\subfloat[]
	{\includegraphics[width=4cm]{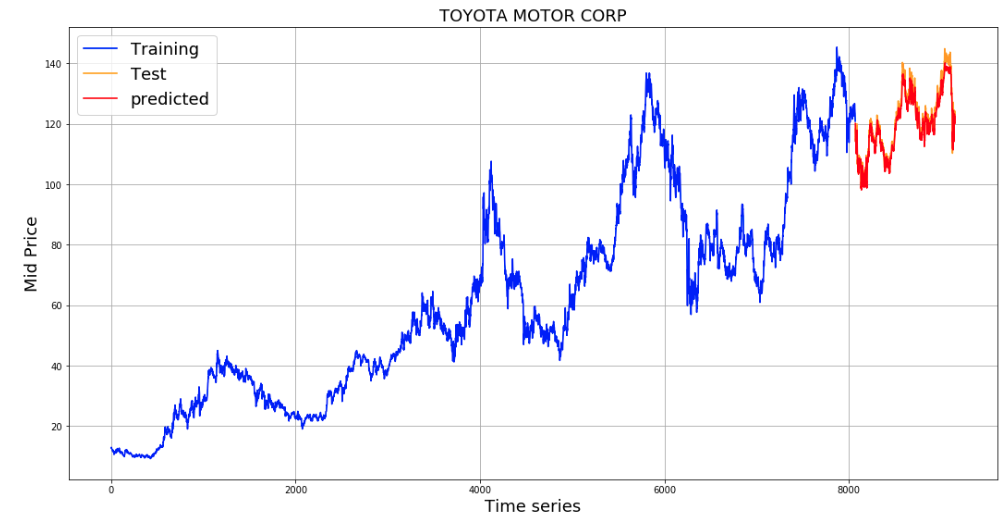}} 
	\subfloat[]
	{\includegraphics[width=4cm]{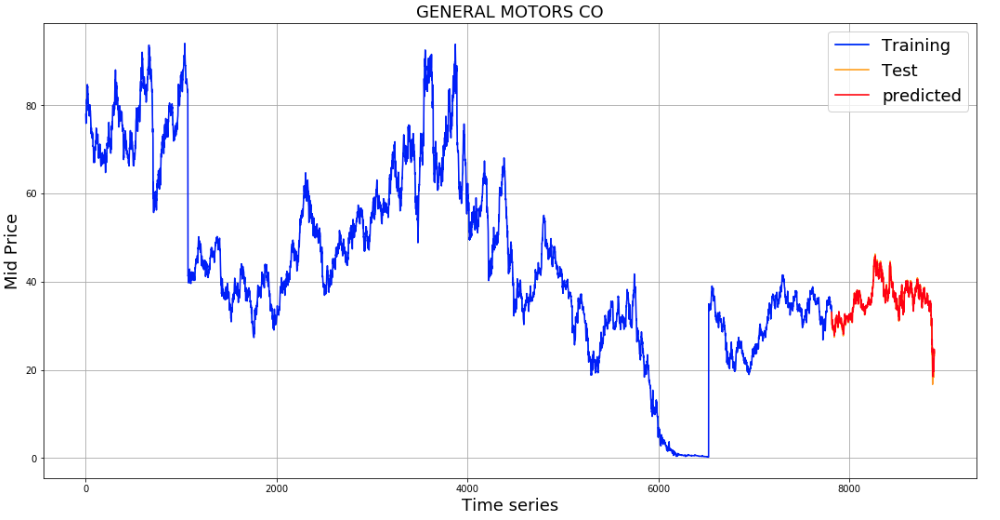}} 
	
	\caption{Stock Price Predictions using Multi-Stack LSTM in case of Ford (a), Tesla (b), Toyota (c) and GM (d)}
	\label{img_grid_LSTM}
\end{figure*}

We trained an ARIMA model, Standard LSTM model, and Custom Loss LSTM model for each company in our dataset. We used the MSE as the training and reporting metric for both LSTM and ARIMA. The MSE obtained on the testing dataset for all four companies is summarized in Table \ref{tab:mse}, and the predictions using ARIMA are plotted in Figure \ref{img_grid_ARIMA}, while those using LSTM are plotted in Figure \ref{img_grid_LSTM}. 
\begin{table}[b]
	\caption{Comparison of MSE between ARIMA and LSTM for all Companies}
	\label{tab:mse}
	\begin{tabular}{lll}
		\toprule
		& \multicolumn{2}{l}{Mean Squared Error} \\
		Company        & ARIMA             & LSTM model           \\ \midrule
		Ford           & 0.025           & 0.1185              \\
		Tesla          & 196.056           & 4624.22             \\
		Toyota         &1.877             & 4.87                \\
		General Motors & 0.34              & 0.413               \\
		\bottomrule
	\end{tabular}
\end{table}
Visually, it is apparent that LSTM and ARIMA make similar predictions on all companies except Tesla. This is borne out from the MSE values in Table \ref{tab:mse}. The LSTM model seems to under-predict the {Tesla}'s stock price during its price boom in the last couple of years. Our belief is that the poor performance of the LSTM model for {Tesla}'s stock price prediction could be due to a much smaller size of its training data as compared to other companies. However, as discussed before, the lower MSE on its own does not guarantee higher profit. To validate the usefulness of our predictions, we will describe our results in Section \ref{subsec:pob} using the trading bot articulated in Section \ref{sec:pob}.

\subsection{Portfolio Optimization Bot} \label{subsec:pob}
\begin{figure}[b]
	\centering
	\includegraphics[width=8cm]{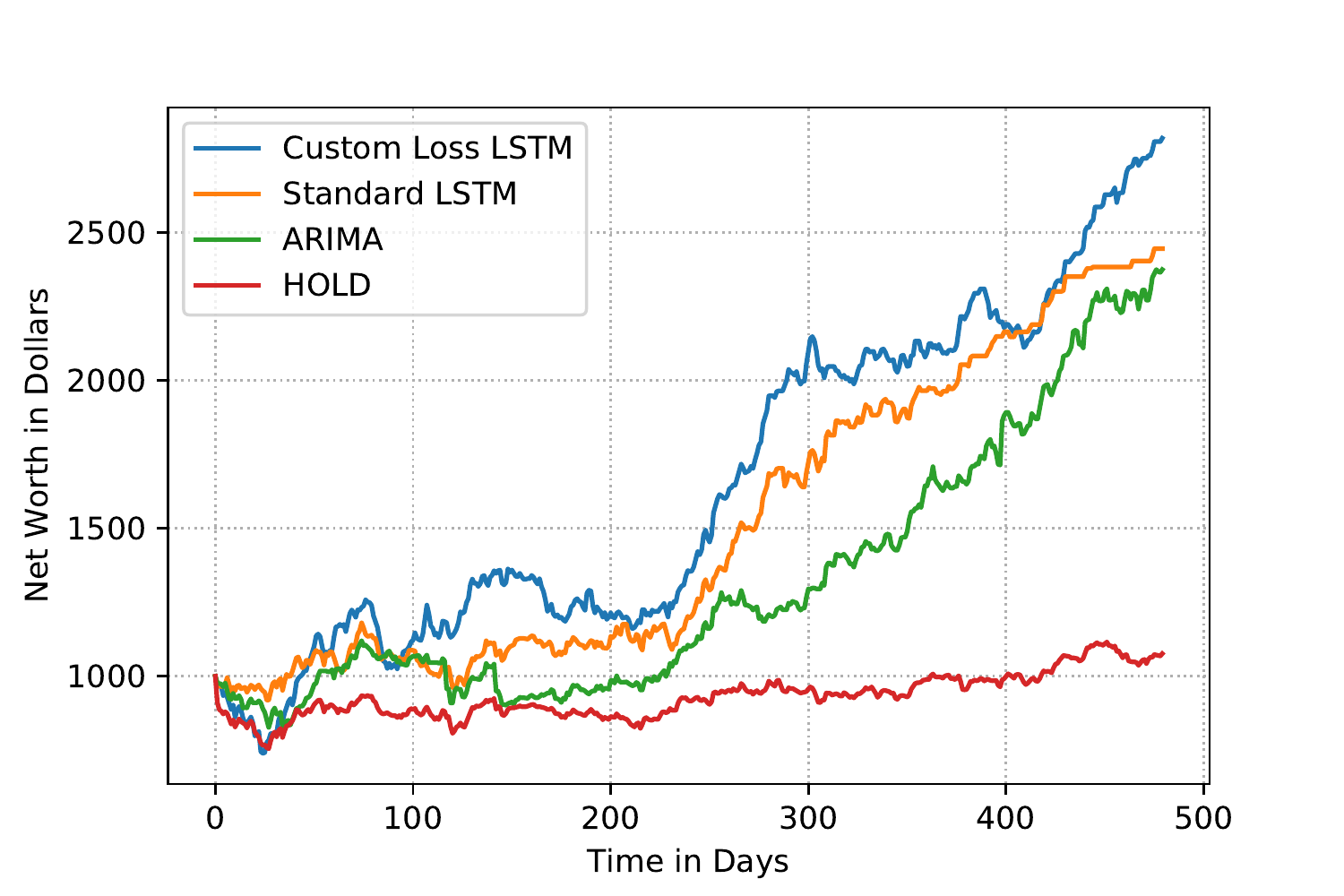}
	\caption{Line Plot Comparing the Performance of ARIMA, Standard LSTM and Custom Loss LSTM with the HOLD operation}
	\label{fig:trading_bot}
\end{figure}
\begin{figure}[b]
	\centering
	\includegraphics[width=8cm]{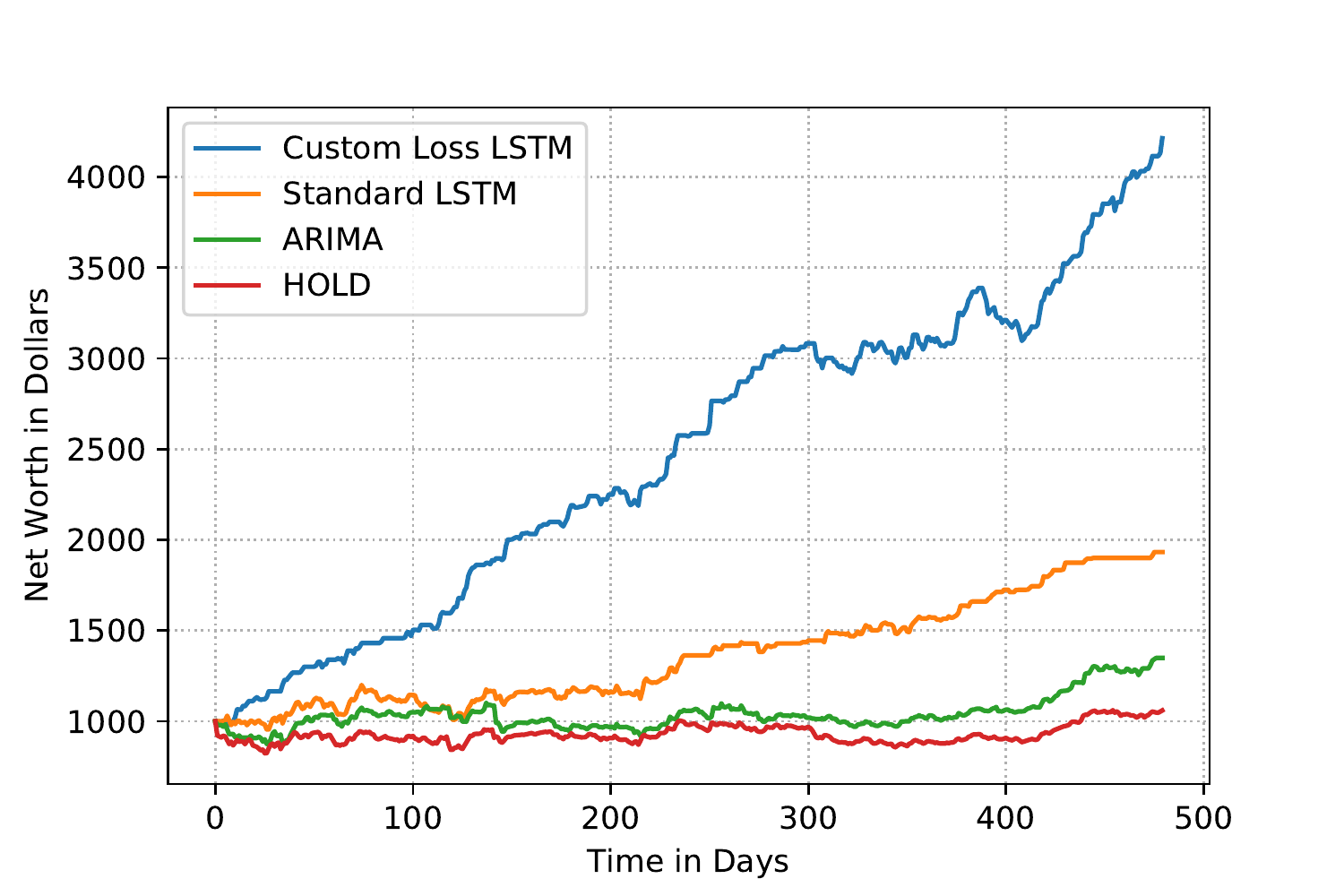}
	\caption{Excluding the Highly Volatile Tesla, Custom Loss LSTM Outperforms all other Methods}
	\label{fig:trading_bot_no_tesla}
\end{figure}
Our portfolio optimization bot is initially given \$1000.00 to invest over a time frame of about 500 days. Figure \ref{fig:trading_bot} shows the portfolio value as a function of time when the bot is used to trade all four companies, with curves corresponding to the Custom Loss LSTM, Regular LSTM, and ARIMA model. The obtained curves have been compared with the scenario when the HOLD operation is applied to the portfolio. It is seen that both LSTM models perform better than the ARIMA model as well as the HOLD operation. That said, we felt the LSTM model performances were slightly underwhelming compared to ARIMA. We deduced that removing Tesla, whose LSTM predictions were poor due to its high volatility, could improve performance. Figure \ref{fig:trading_bot_no_tesla} shows the bot performance without Tesla. As one can see, the Custom Loss LSTM model's performance became significantly better than the other models, more than quadrupling the initial investment. That said, the Standard LSTM and ARIMA models both under-performed as compared to Figure \ref{fig:trading_bot}, though this too can be attributed to removing Tesla, as more volatile stocks can lead to both greater gains and losses. It seems those models had a significant portion of their profit associated with the Tesla stock.
\section{Extensions}
\subsection{Adding Correlated Indicators}
\subsubsection{Adding Analyst Call}
Analyst calls are expert predictions for Earning Per Share (EPS) of a company and show weak correlation with the company’s stock price. These calls are placed as quarterly, yearly, bi-yearly, etc. forecasts. We decided to augment our stock price data with the quarterly forecast analyst call data to see if it could improve the MSE in multi-step LSTM prediction. Since these calls are only reported every few days, we smoothed the data using a forward fill exponential moving average with a window length of 12 to make it useful for the model.

\subsubsection{Bidirectional LSTM with Attention}
\begin{figure}[b]
	\includegraphics[width=7cm,height=6.65cm]{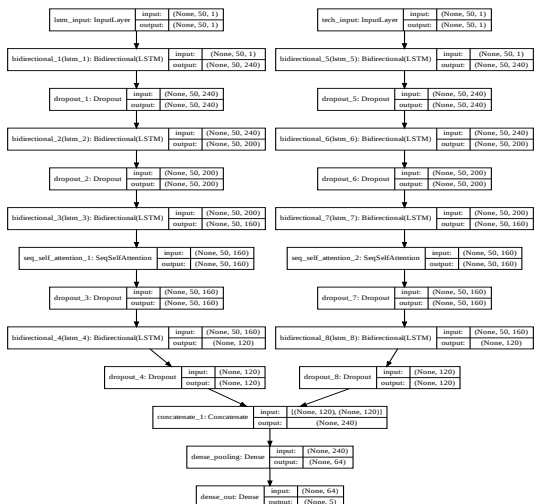}
	\caption{Bidirectional LSTM with Attention Architecture}
	\label{fig:model_attention}
\end{figure}

As shown in Figure \ref{fig:model_attention}, the stock prices are passed through the multi-stack LSTM on the left and the analyst calls through the LSTM on the right. Each LSTM stream consists of 4 Bidirectional LSTM units with a single Attention layer. The two streams of LSTM are then concatenated and sent through a dense layer. Bidirectional LSTMs enable us to fit the data better by incorporating past and future dependencies during training, while Attention selectively chooses which inputs to weigh more given all the past inputs. Many articles \cite{li2018stock,kim2019forecasting,qiu2020forecasting} have advocated usage of Attention to improve prediction performance. We hoped this would help us capture hidden trends in the data since there is an intrinsic lag between an analyst forecast and what actually happens with the stock price. These changes caused reduction in MSE from 0.20 to 0.19 using the Ford stock prices. However, it did not work well on the other datasets. This is likely due to an imbalance in number of parameters versus number of data points, implying that we could not guarantee convergence of our model. Our results suggest adding correlated indicators has the potential to improve the model's performance, but it is highly data dependent. 

\subsection{Varying Training Window and Prediction Length}

\begin{figure}
	\includegraphics[width=7cm]{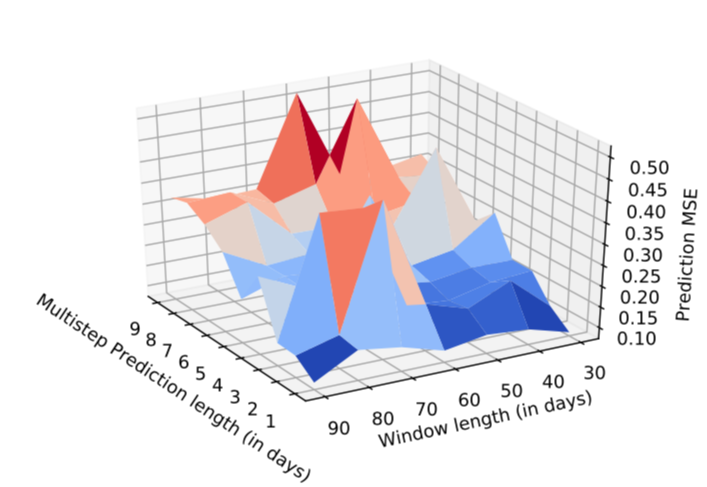}
	\caption{Varying Training Window and Prediction Length  }
	\label{3d_plot}
\end{figure}

We varied the window length, which is the number of past stock prices used for prediction, from 30 to 90 days, and the number of future days predicted by the regular LSTM from 1 to 9 off the same window. The results for Ford are shown as a heat map in Figure \ref{3d_plot}. This is a data driven approach to finding the optimal hyperparameters. As expected, lowering the number of future days predicted results in a lower MSE. However, what would the optimal window length be if one wanted to predict multiple days in advance? The main goal is to find a sweet spot for the combination of the window length and number of future days predicted that has the lowest MSE. Repeating this procedure for all companies to find the global minimum for MSE would provide optimal hyperparameters for our LSTM models to perform best in the portfolio optimization bot.

\section{Conclusion {\&} Future directions}
Prediction of stock prices in the future is an attractive application of modern machine learning algorithms. Both ARIMA and LSTM showed comparable accuracy for stock price predictions on majority of the data, though the LSTM fares poorly on highly volatile stocks, and ARIMA outperforms it for our datasets. None of the prior studies defined a trading strategy to investigate the profit one could earn using their predictions. We developed a portfolio optimization bot using convex optimization techniques, which was exploited to automate the process of investing in the stock market end-to-end. Moreover, LSTMs have a more flexible training procedure that we modified to indirectly maximize the profit. To incorporate correlated indicators such as analyst calls, we extended the regular LSTM model to a double-stream Bidirectional LSTM architecture with Attention. Data driven optimization of window length and multi-step length prediction are two of the tasks that seem viable in the future for the improvement of our predictions and, in turn, our portfolio manager. Moreover, sparse auto-encoders with 1-D residual convolutional networks could be used to denoise the data to improve the performance of the Bidirectional LSTM.

\bibliographystyle{ACM-Reference-Format}
\bibliography{references}
\end{document}